\documentclass[conference]{IEEEtran}
\usepackage{cite}
\usepackage{amsmath,amssymb,amsfonts}
\usepackage{graphicx}
\usepackage{xcolor}
\usepackage{arabtex}
\usepackage{enumitem}
\usepackage{utf8}
\usepackage{caption} 
\usepackage{soul}   
\usepackage{multirow}
\usepackage{url} 

\setcode{utf8}
\begin{document}
\title{ANER: Arabic and Arabizi Named Entity Recognition using Transformer-Based Approach}

\author{

\IEEEauthorblockN{
Abdelrahman "Boda" Sadallah}
\IEEEauthorblockA{\textit{Computer and Systems Engineering
} \\
\textit{Ain Shams University}\\
Cairo, Egypt \\
bodasadallah@gmail.com}
\and

\IEEEauthorblockN{
Omar Ahmed}
\IEEEauthorblockA{\textit{Computer and Systems Engineering} \\
\textit{Ain Shams University}\\
Cairo, Egypt \\
key.omar10@gmail.com
}
\and

\IEEEauthorblockN{
Shimaa Mohamed
}
\IEEEauthorblockA{\textit{Computer and Systems Engineering} \\
\textit{Ain Shams University}\\
Cairo, Egypt \\
shimaamohamed18286@gmail.com
}

\and
\IEEEauthorblockN{
Omar Hatem
}
\IEEEauthorblockA{\textit{Computer and Systems Engineering} \\
\textit{Ain Shams University}\\
Cairo, Egypt \\
omarhatem221@gmail.com
}
\and
\IEEEauthorblockN{
Doaa Hesham
}
\IEEEauthorblockA{\textit{Computer and Systems Engineering} \\
\textit{Ain Shams University}\\
Cairo, Egypt \\
doaahesham1399@gmail.com
}
\and
\IEEEauthorblockN{
Ahmed H. Yousef}
\IEEEauthorblockA{\textit{Computer and Systems Engineering} \\
\textit{Ain Shams University}\\
Cairo, Egypt \\
Ahmed.Yousef@EUI.edu.eg}}

\IEEEoverridecommandlockouts
\IEEEpubid{\makebox[\columnwidth]{979-8-3503-3556-9/23/\$31.00 ©2023 IEEE \hfill}
\hspace{\columnsep}\makebox[\columnwidth]{ }}
\maketitle
\IEEEpubidadjcol




\begin{abstract}
One of the main tasks of Natural Language Processing (NLP), is Named Entity Recognition (NER). It is used in many applications and also can be used as an intermediate step for other tasks. We present ANER, a web-based named entity recognizer for the Arabic, and Arabizi languages. The model is built upon BERT, which is a transformer-based encoder. It can recognize 50 different entity classes, covering various fields. We trained our model on the WikiFANE\_Gold dataset which consists of Wikipedia articles. We achieved an F1 score of 88.7\%, which beats CAMeL Tools' F1 score of 83\% on the ANERcorp dataset, which has only 4 classes. We also got an F1 score of 77.7\% on the NewsFANE\_Gold dataset which contains out-of-domain data from News articles.
The system is deployed on a user-friendly web interface that accepts users' inputs in Arabic, or Arabizi. It allows users to explore the entities in the text by highlighting them. It can also direct users to get information about entities through Wikipedia directly. We added the ability to do NER using our model, or CAMeL Tools' model through our website. ANER is publicly accessible at \url{http://www.aner.online}. We also deployed our model on HuggingFace at https://huggingface.co/boda/ANER, to allow developers to test and use it.

\end{abstract}


\section{Introduction} The importance of Named Entity Recognition (NER)
comes from the fact that it is used in many NLP tasks such as question answering (QA), machine translation (MT), and sentiment analysis (SA). The NER task consists of two main steps: (1) extracting the named entities in the text, and (2) classifying these entities into different groups. The main classes for the NER task are Person, Organization, Location, and Miscellaneous.\cite{tjong-kim-sang-de-meulder-2003-introduction} However, these can be segregated into many domain-specific classes.

We aim to advance the progress in the NER task for Arabic by introducing ANER. It is a NER model for the Arabic language, built upon AraBERT\cite{antoun-etal-2020-arabert}. AraBERT is a pre-trained BERT\cite{devlin-etal-2019-bert} model for Arabic. BERT models consist of stacked transformer\cite{Vaswani2017} bidirectional encoder blocks. We trained the model on the WikiFANE\_Gold\cite{alotaibi-lee-2014-hybrid} dataset which is an Arabic Named Entity corpus of Wikipedia-based articles. We can recognize 50 classes such as Nation, Media, Software, Educational, Artist, etc. We deployed our system online as a web interface, to make it available to as many people as possible, and added support for the Arabizi language, as many people use it as their main language for communication. Arabizi is a method of typing Arabic text using English letters and numbers. For example, the Arabic word: \RL{مُعلِّمْ} is written as \textbf{Mo3allem} in Arabizi. \\
As part of our contribution to the community, we open-sourced all of our training scripts\footnote{\url{https://github.com/BodaSadalla98/Arabic-NER}} and the website source code on GitHub\footnote{\url{https://github.com/BodaSadalla98/aner}}.
Our contributions are as follows:
\begin{itemize}

\item We support 50 different entities, spanning a wide range of classes.

\item Added support for the Arabizi language, as it is being used by more and more people.

\item Deployed the system to an online web interface\footnote{\url{http://www.aner.online}}, to make it easily accessible from anywhere.
\item ANER model is deployed on the HuggingFace\footnote{\url{https://huggingface.co/boda/ANER}} model repository.
\item Open-sourced our training scripts and the code for the User Interface.
\end{itemize}

\section{Motivation}
The Arabic language is a complex language. This resulted in the difficulty of doing many of the NLP tasks in it. Some of the main reasons are: 
\begin{itemize}
\item The rich inflectional and cliticization morphology system in Arabic. For example, the Modern Standard Arabic (MSA) word, \RL{فَسَيَكْفِيكَهُمُ} fasayakfīkahumu ‘So will suffice you against them’ has two proclitics (conjunction and a tense marker), and two enclitics (objects).
\item The diacritization writing system in Arabic is optional. This creates a high degree of ambiguity. As shown in Table \ref{tab:ambiguity_arabic}, one Arabic word can have many different meanings depending on its diacritization. The Standard Arabic Morphological Analyzer\cite{Maamouri2010} produces on average 12 different analyses per MSA word.

\item  Almost every Arab country has its own unique dialect. On top of that, there can be multiple different dialects within the same country. These dialects significantly diverge from the MSA which is the formal Arabic language. Using NLP tools that are built for MSA, to process any of these dialectal texts, results in low performance. For example, one of the top morphological analyzers for MSA covers only 60\% of Levantine Arabic verb forms\cite{habash-etal-2012-morphological}.

\item Generally, Deep Learning models rely on a huge amount of training data, which is not easily available for Arabic.
\item Concatenation of affixes and clitics on stems, make Arabic one of the richest languages in morphology\cite{AlSughaiyer2004}.
\end{itemize}

The NER task for Arabic has some difficulties in itself:
\begin{itemize}
\item  Words can be mapped to different entities according to the context. For example, the word \RL{الزمالك} can be labeled as an "Organization," or a "Location."
\item The absence of capitalization in Arabic adds another layer of ambiguity and makes the task harder.
\end{itemize}

There is a need for tools and solutions that tackle the main NLP tasks in Arabic. As well as, taking into consideration the unique nature of Arabic and its challenges.

{\renewcommand{\arraystretch}{1.1}
\begin{table}[h!]
    \centering
    \begin{tabular}{c  c c}
    \hline 
    Arabic word & Transliteration & English translation\\
    \hline
    \RL{عَلِمَ}  & Alema & Knew\\
    \hline
    
    \RL{عَلَّمَ}  & Allama& Taught\\
    \hline
    
    \RL{عِلم} &Elm & Knowledge\\
    \hline
    
    \RL{عَلَم} & Alam & Flag\\
    \hline
     \end{tabular}
    \caption{Arabic words can have different meanings depending on diacritics.}
    \label{tab:ambiguity_arabic}

\end{table}}

\begin{figure}[htp]
    \centering
    \includegraphics[width=8cm]{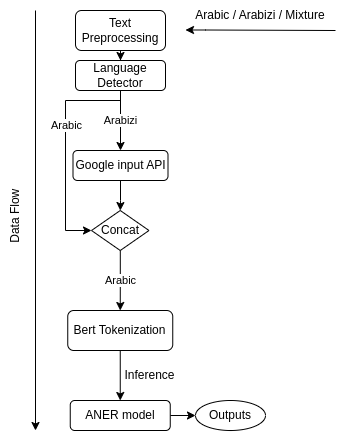}
    \caption{System Pipeline.}
    \label{figure:deployment}
\end{figure}

\section{related work}
There exist many libraries and solutions that support different Arabic NLP tasks, such as diacritization, sentiment analysis, morphological analysis, and NER. Some of those libraries support Arabic NER with different features, accuracy, speed, and usability. We will discuss some of them. CAMeL Tools\cite{obeid-etal-2020-camel}, Farasa\cite{abdelali-etal-2016-farasa}, and Madamira\cite{pasha-etal-2014-madamira} are the most effective solutions that support NLP tasks for the Arabic language. We tested all these solutions to find their pros and limitations.

\begin{itemize}

\item Farasa is a famous Arabic NLP package\footnote{\url{https://farasa.qcri.org/}}. It has a Web API as well as, a user interface demos. They support a variety of tasks such as segmentation, spell-checking, part-of-speech tagging, lemmatization, diacritization, and Arabic NER. The main problem with using it for NER is that the user interface demo highlights the entities only without classifying them.

\item CAMeL Tools is a well-known open-source library that supports multiple language processing tasks, one of which is Arabic NER. They used the pre-trained AraBERT model as we do, and trained it on the ANERcorp dataset\cite{Benajiba2007}, which is Arabic Named Entity Corpus with around 150K tokens. They applied some changes\footnote{\url{https://camel.abudhabi.nyu.edu/ANERcorp}} to the original dataset, including making new splits for train, and test. The library has some limitations, as (1) it doesn't have a user interface to utilize its models, and (2) its NER model supports 4 entity classes only.
\end{itemize}
The current solutions have limitations that need to be addressed. These can be summarized in a few major points:
\begin{itemize}
  \item  Online, user-friendly solutions (CLI, website, demos, desktop apps, etc.).
  \item Support of many entity classes other than the four main classes (e.g., Athlete, and Population-Center).
  \item Reliable models with high accuracy.
  \item Providing more information about each entity. For example, highlighting the entity and giving a link to more information about it.
  \item Support of Arabizi.
  \item Support of different Arabic dialects.
\end{itemize}
In our system, we tried to address all these points.

\begin{figure}[htp]
    \centering
    \includegraphics[width=8cm]{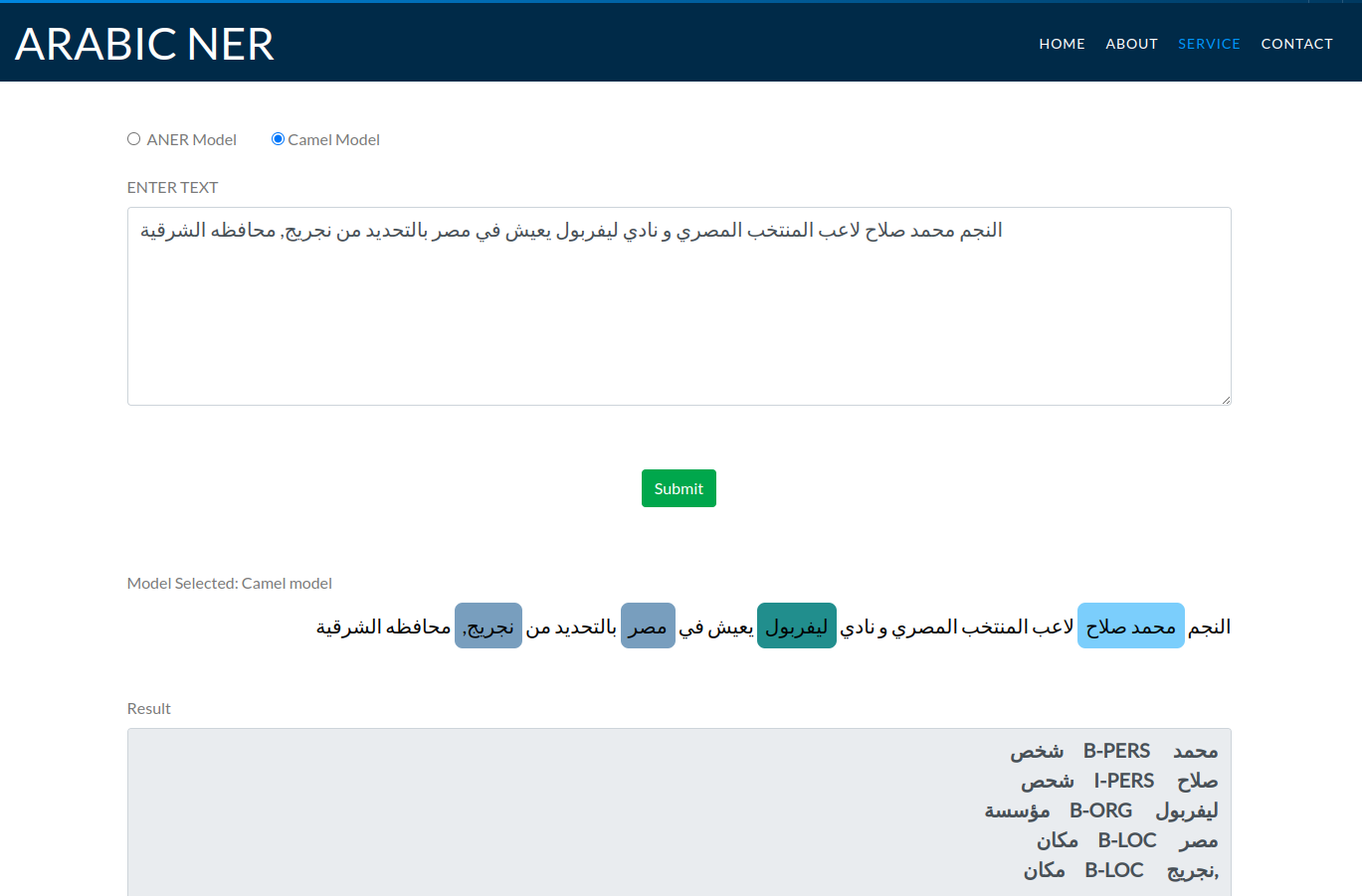}
    \caption{The ANER interface with an Arabic example using CAMeL Tools model.}
    \label{fig:arabic_inf_2}
\end{figure}

\section{METHODOLOGY}

\subsection{DATASETS} 
We used a publicly available dataset from King Abdulaziz University. It's a high-quality Arabic NER corpus\cite{alotaibi-lee-2014-hybrid}. We used the manually gold-standard corpus which has two parts:
\begin{itemize}
    \item WikiFANE\_Gold: it is a manually annotated dataset, which has around 500k tokens. The data consists of Wikipedia-based articles and is annotated into 50 entity classes.
    \item NewsFANE\_Gold: a dataset contains 170k tokens. It has the same textual data as ANERcorp but was annotated, with 50 entity classes instead of 4.
\end{itemize}
We also used the ANERcorp dataset\cite{Benajiba2007}. It's a manually annotated corpus of 150k tokens, coming from 316 articles selected from news wire. It's annotated using the standard CoNLL format\cite{tjong-kim-sang-de-meulder-2003-introduction}, and with the 4 main NER classes mentioned in the introduction.

We trained our model on the WikiFANE\_Gold dataset using 80\% training, 10\% evaluation, and 10\% testing splits. We used NewsFANE\_Gold as a real test of out-of-distribution data. We also trained another instance on the ANERcorp dataset using CAMeL Tools' data splits\cite{obeid-etal-2020-camel}.

\subsection {DATA PREPROCESSING }
\subsubsection{BERT Specific Preprocessing}
Bert requires some specific preprocessing for the data to be ready for training. One of these specifications is that all training samples must have the same length. In order to choose the best sequence length, we performed a detailed analysis of the dataset. We found that most of the sentences do not exceed the limit of 256 tokens, so we chose that as our sequence length.
\subsubsection{Tokenization}
We used the AraBERT tokenizer, which has a vocabulary size of 64k tokens including 4k unused tokens, allowing the addition of new tokens for more customization. Bert tokenizers are sub-word-based tokenizers\cite{kudo-richardson-2018-sentencepiece} that split the words into tokens and substitute every token with a unique ID. For token classification tasks such as NER, we have to make a choice for how to tokenize words that correspond to entities. Usually, these words would get split into multiple tokens by the tokenizer, and then we can choose to (1) label all the word's tokens with the correct entity-label, or just give the entity-label to the first token, and label the rest of the tokens with [PAD]. Table \ref{tab:labeling_approaches} shows an example of this. We tried tokenizing the labels using the two approaches but didn't find any big differences in performance.

{\renewcommand{\arraystretch}{1.2}
\begin{table}[h!]
    \centering
    \begin{tabular}{c c c c}
    \hline
    Word & \multicolumn{3}{c}{\RL{القاهرة}}\\
         \hline 
         Sub-words &  \RL{\#\#ة} &  \RL{\#\#قاهر}  & \RL{ ال}\\
         \hline 
        Corresponding tokens & 10 &457& 1502 \\
        \hline
        Approach 1 Labels  & B-Loc & B-Loc & B-Loc  \\
        \hline
        Approach 2 Labels & O & O & B-Loc  \\
        \hline
    \end{tabular}
    \caption{Two token classification labeling approaches.}
    \label{tab:labeling_approaches}
\end{table}}

\subsection{AraBERT Model} 
AraBERT\cite{antoun-etal-2020-arabert} is basically a BERT-based model, that is trained on a lot of MSA data. It is trained on articles that are manually scrapped from Arabic news websites, a massive corpus of 1.5 billion words\cite{ElKhair201615BW}, and an open-source Arabic news corpus\cite{zeroual-etal-2019-osian}. The datasets sum up to around 70 million sentences and around 24GB of text. The basic model architecture consists of 12 encoder blocks. Each of which, has 12 attention heads, and a hidden size of 768. This gives a total of 110M parameters. As for the maximum sequence length, AraBERT models can take inputs of sizes up to 512 tokens. The pre-training process of AraBERT is done by applying two tasks\cite{devlin-etal-2019-bert}: (1) a masked language modeling task (MLM). They randomly mask 15\% of the input tokens, and train the model to predict them.  (2) They also used the next sentence prediction (NSP) task, where they enter two sentences into the model and predict whether the second sentence comes after the first in the original data. This task helps the model to capture the long-term relations across the sentences.
\subsection{Fine-Tuned Model} 
AraBERT provides many models with different sizes. We used the AraBERTv0.2-base pre-trained model from HuggingFace's open-source Transformers library\cite{wolf-etal-2020-transformers} as our base model. To fine-tune the model on the NER task, we added a Fully Connected (FC) layer that has 102 outputs (101 class labels, and one non-entity label). We followed that with a cross-entropy loss layer to calculate the loss and back-propagate the error signal. Finally, the node that has the highest probability is taken as the predicted class of this token.

\subsection{Training setup}
We trained our models using the free service of Google Colaboratory\footnote{\url{https://colab.research.google.com/}}. We used 15GB of GPU memory and trained the model for 10 epochs. For the training learning rate, we did many experiments using different values and achieved the best results with learning rate = 5x10\textsuperscript{-4}.

\begin{figure}[htp]
    \centering
    \includegraphics[width=8cm]{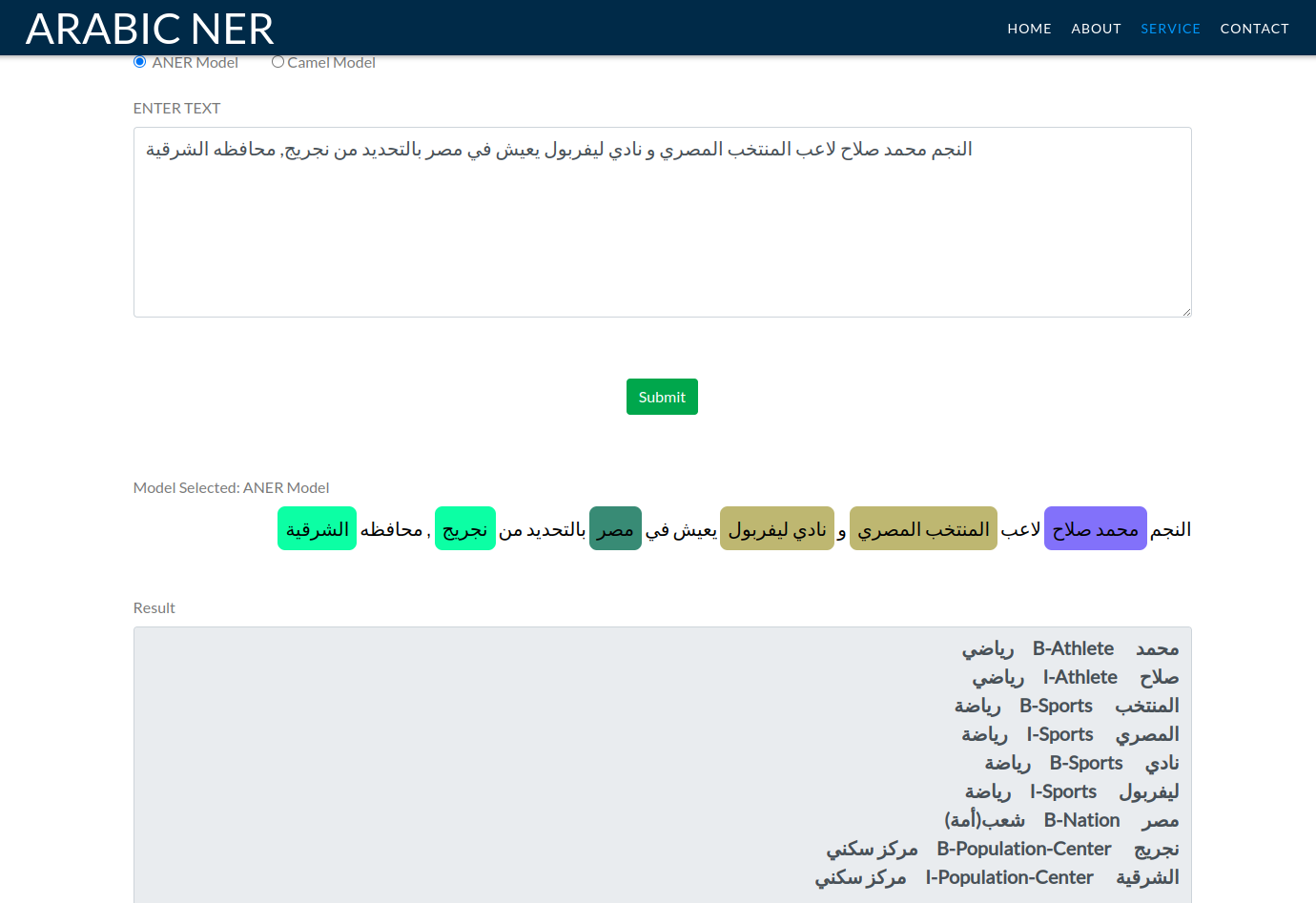}
    \caption{The ANER interface with the same Arabic example using our ANER model. In this example, more entities are recognized, and the entities are more specific.}
    \label{fig:arabic_inf_1}
\end{figure}

\section{System Design and Implementation}
\subsection{Design motivation and pipeline}
\subsubsection{Design motivation}
We wanted a system that has easy access, and that is easy to use for people that are interested in Arabic NER. That's why we created an application that supports input text in Arabic, Arabizi, or a mixture of both. To add the ability to gain more information about the recognized entities, we highlighted the entities and made them clickable, so that the user can be redirected to the entity's page on Wikipedia.
\subsubsection{Pipeline}
To achieve the system requirements, we designed a pipeline that the data go through till we get the output entities. The process stages are presented in Figure \ref{figure:deployment}. It starts with the user entering a sentence, which goes through a preprocessing step, followed by a language detector that is used to separate the Arabizi text and enter it into Google input tools API\footnote{\url{https://www.google.com/inputtools}}. It then gets transliterated back into Arabic. Finally, it gets concatenated with the original Arabic words. At this stage, the input sentence contains only Arabic characters. Lastly, we apply Bert-specific preprocessing before doing model inference and getting the outputs.

\subsection{Implementation}

\subsubsection{ Text Preprocessing}
This is a simple preprocessing step that cleans the input text by removing: new lines, spaces, and emojis so that it could be transliterated using Google Input API.
\subsubsection{Language Detector}
This module is associated with classifying every word of the input sentence. If the word is Arabizi, it will be entered into the transliteration module. Otherwise, it will go to the concatenation step directly. 
\subsubsection{Google Input API}
we searched for tools that can be used to transliterate Arabizi into Arabic. We found a few solutions that do this. One of the most popular services that exist is Yamli\footnote{\url{https://www.yamli.com}}. We could not rely on Yamli, as it suffers from a few limitations, most importantly is the fact that they don't provide an API to use their services. On the other hand, Google has a free easy-to-use API, as part of its Google services. We used it to transliterate Arabizi words into Arabic words. We enter the Arabizi data word by word into the API and get the corresponding Arabic output.
\subsubsection{Bert Tokenization}
By this stage, the input sequence would contain only Arabic words. We apply the processing steps required by Bert models, starting from unifying all sentences to have the maximum sequence length. After that, we apply the sub-word tokenization.

\subsubsection{Deployment}
\begin{figure}[htp]
    \centering
    \includegraphics[width=8cm]{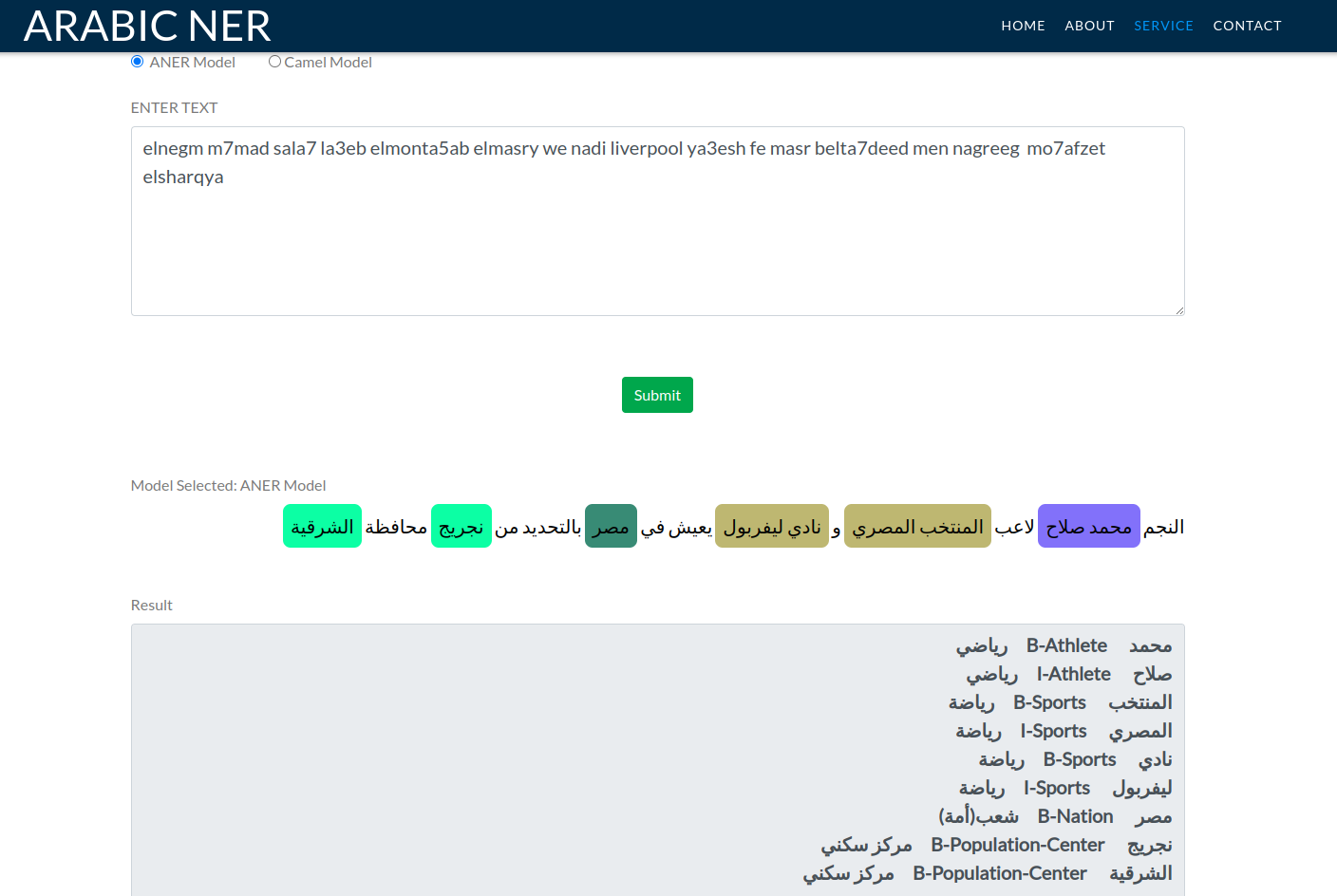}
    \caption{The ANER interface with the same example, but written in Arabizi.}
    \label{fig:arabizi_inf}
\end{figure}

We packaged our system in an easy-to-use User Interface that is available online. The user can write a sentence either in Arabic, Arabizi, or both, and get the recognized named entities. The output is also displayed with entities of the same class highlighted with a unique color. The deployment process is mainly divided into two main modules:

\begin{itemize}
\item Interface: the front-end part is implemented using HTML, CSS, and JavaScript. The UI module contains a simple text box where you can enter a sentence in Arabic, Arabizi, or both, and get the recognized entities highlighted. It contains a check box, to choose between our model and the CAMeL Tools model.
 In figure \ref{fig:arabic_inf_2}, we  show an example of a sentence entered in Arabic and its output using the CAMeL Tools model.
Figure \ref{fig:arabic_inf_1} shows the output of the same sentence using our ANER model.
Figure \ref{fig:arabizi_inf} shows the same example as above but written in Arabizi. 
\item Back-end: as for the back-end module, we decided to use Flask\footnote{\url{https://palletsprojects.com/p/flask}}, as it easily connects the two deployment modules, and  has multiple modules that help developers write applications without worrying about thread or protocol management details.
\end{itemize}


\section{ RESULTS AND DISCUSSION} 
We trained our model on WikiFANE\_Gold which is our main dataset. As we have an imbalanced set of samples per class, we relied on the recall, precision, and F1 scores to evaluate our model instead of using accuracy. We also trained another instance of our model on the ANERcorp dataset, to compare our model with CAMeL Tools’ model. To get reliable test results, we tested our system against two test sets. The first is the test set portion of our main training dataset. We also tested our system with the NewsFANE\_Gold dataset, which contains data from a different domain than our training data.


\subsection{ Results}
\subsubsection{Testing with ANERcorp test set} 
We trained our model on the ANERcorp dataset using CAMeL Tools' splits. In table \ref{tab:anercorp_results}, we compare our results on the test set split with the results obtained by CAMeL Tools in their paper\cite{obeid-etal-2020-camel}.

{\renewcommand{\arraystretch}{1.5}
\begin{table}[h!]
    \centering
    \begin{tabular}{|c | c|c |} 
    \hline
        Metric & Our model & CAMeL Tools  \\
         \hline 
        Recall & \textbf{82.0} & 81.0\\
        \hline
        Precision & \textbf{84.7} & 84.0 \\
        \hline
        F1 & \textbf{83.3} & 83.0\\
        \hline

    \end{tabular}
    \caption{Our ANER model results compared to CAMeL Tools on the ANERcorp dataset.}
    \label{tab:anercorp_results}
\end{table}}



\subsubsection{Testing with WikiFANE\_Gold and NewsFANE\_Gold datasets}
We trained another instance of our model on the WikiFANE\_Gold dataset, which is more than triple the size of ANERcorp. We achieved around 5\% improvement across all evaluation metrics. To further test our model, we tested it across the NewsFANE\_Gold dataset which contains out-of-domain data. In table \ref{tab:aner_results} we compare our results on the two datasets.


{\renewcommand{\arraystretch}{1.5}
\begin{table}[h!]
    \centering
    \begin{tabular}{|c | c|c |} 
    \hline
        Metric & WikiFANE\_Gold & NewsFANE\_Gold  \\
         \hline 
        Recall & 87.0 & 78.1 \\
        \hline
        Precision &  90.5 & 77.4  \\
        \hline
        F1 & 88.7 & 77.7 \\
        \hline

    \end{tabular}
    \caption{Test results of our ANER model on WikiFANE\_Gold and NewsFANE\_Gold datasets.}
    \label{tab:aner_results}
\end{table}}

\subsection{Discussion} 
We achieved almost the same results as CAMeL Tools on the same dataset (ANERcorp). We got an improvement in the performance on our test set of more than 5\%, and we deduce that it's because our training dataset is bigger than ANERcorp. There's also a drop in performance of around 10\% on the NewsFANE\_Gold dataset, and we suspect that it's because the data comes from a different domain than our training dataset.

\section{CONCLUSION AND FUTURE WORK}
\subsection{Conclusion}

We present ANER, a Named Entity Recognizer for Arabic. It is trained on a relatively large dataset (500k Tokens). In addition, we support 50 different entity classes, in contrast to only 4 entities supported by the other frameworks. We also added support for Arabizi, which is becoming more popular. We deployed our system online, in a user-friendly UI, so that it becomes accessible to anyone.

\subsection{Limitations}
We acknowledge that our model is only trained on Modern Standard Arabic data and that it can give poor outputs when challenged with heavy-dialect-specific inputs. We are also limited by the quality of the Arabizi transliterating to give correct results for the Arabizi inputs.
\subsection{Future Work} 
We believe that this is just a step in the Arabic NER progress and that there is still much room for improvement. Some future work includes:
\begin{itemize}
    \item Train on a bigger dataset, that contains multi-domain data, so that we can decrease the generalization error.
    \item Support different Arabic dialects.
    \item  Try different pre-trained backbones other than Bert. For example, MARBERT\cite{abdul-mageed-etal-2021-arbert}.
    \item Use our ANER model in other Arabic NLP tasks (e.g., Question Answering task).

\end{itemize}

\section{ACKNOWLEDGMENT}
We want to begin by thanking our supervisors Prof. Ahmed H. Yousef and Eng. Sara Abd Elaziz for their dedicated supervision, and continuous support during our graduation project. We are extremely grateful to the AraBERT team for their work in enriching the Arabic NLP community with open-source models, and to the people at King Abdulaziz University for making such high-quality datasets publicly available. Lastly, we would like to thank Eng. Baraa Magdy for her efforts in reviewing this paper.

\bibliographystyle{IEEEtran}
\bibliography{aner_bib.bib}

\vspace{12pt}

\end{document}